\pdfoutput=1

\documentclass[11pt]{article}

\usepackage[]{acl}

\usepackage{times}
\usepackage{latexsym}

\usepackage[T1]{fontenc}

\usepackage[utf8]{inputenc}

\usepackage{microtype}

\usepackage{natbib}
\usepackage{microtype}
\usepackage{booktabs, graphicx}
\usepackage{xcolor}
\usepackage{forest}
\usepackage{algorithm}
\usepackage{algpseudocode}

\usepackage{graphicx}
\usepackage{amsmath}   
\usepackage{amssymb}
\usepackage{xspace}

\newcommand{\added}[1]{#1}
\newcommand{\addedB}[1]{#1}

\newcommand{\regex}[1]{\ensuremath{\mathrm{#1}}}
\newcommand{\str}[1]{\ensuremath{\mathsf{#1}}}
\newcommand\acc{\mathrm{acc}}
\newcommand\perf{\mathrm{perf}}
\newcommand\perfbf{\mathrm{\mathbf{perf}}}
\DeclareMathOperator*{\avg}{avg}
\newcommand{\instr}{r}
\newcommand{\instrlang}{\mathcal{R}}
\newcommand{\rlang}{\text{r-language}\xspace}
\newcommand{\rlangs}{\text{r-languages}\xspace}

\newcommand{\eat}[1]{}

\usepackage{quoting}

\newenvironment{ite}{                     
     \parskip 0cm \begin{itemize} \parskip 0cm \parsep 0cm \itemsep 0cm \topsep 0cm}{
        \end{itemize}} 

\newcommand{\ie}{i.e.,\xspace}
\newcommand{\eg}{e.g.,\xspace}

\title{What Makes Instruction Learning Hard?\\ An Investigation and a New Challenge in a Synthetic Environment}

\author{
    Matthew Finlayson \hspace{3ex}
    Kyle Richardson \hspace{3ex} 
    Ashish Sabharwal \hspace{3ex}
    Peter Clark \\
    Allen Institute for AI, Seattle, WA \\
    \texttt{\{matthewf,kyler,ashishs,peterc\}@allenai.org}
}

\begin{document}

\maketitle

\begin{abstract}
The instruction learning paradigm---where a model learns to perform new tasks from task descriptions alone---has become popular in general-purpose model research.
The capabilities of large transformer models as instruction learners, however, remain poorly understood.
We use a controlled synthetic environment to characterize such capabilities. 
Specifically, we use the task of deciding whether a given string matches a regular expression (viewed as an instruction) to identify properties of tasks, instructions, and instances that make instruction learning challenging. 
For instance, we find that our model, a fine-tuned T5-based text2text transformer, struggles with large regular languages, suggesting that less precise instructions are challenging for models. Additionally, instruction executions that require tracking longer contexts of prior steps are also difficult.
We use our findings to systematically construct a challenging instruction learning dataset, which we call Hard RegSet.
Fine-tuning on Hard RegSet, our large transformer learns to correctly interpret only 65.6\% of test instructions (with at least 90\% accuracy), and 11\%-24\% of the instructions in out-of-distribution generalization settings. 
We propose Hard RegSet as a challenging instruction learning task, and a controlled environment for studying instruction learning.\footnote{Data and code available at \url{https://github.com/allenai/RegSet}}
\end{abstract}


\section{Introduction}

\begin{table}[t]
\centering
    \resizebox{\columnwidth}{!}{
    \begin{tabular}{llc|llc}
    \toprule
         \multicolumn{3}{c}{Train} & \multicolumn{3}{c}{Test}  \\
         \small Instruction & \small Data & \small Result & \small Instruction & \small Data & \small Result \\
         \small (RegEx) & \small (String) & \small (T/F) & \small (RegEx) & \small (String) & \small (T/F) \\
         \midrule
         \regex{a{*}b} & \str{aaa} & F & \regex{(a{*}b){*}} & \str{aabab} & T \\
         \regex{a{*}b} & \str{aab} & T & \regex{(a{*}b){*}} & \str{aba} & F \\
         \regex{(ab){*}} & \str{aab} & F & \regex{(a{*}b){*}} & \str{aab} & T \\
         \regex{(ab){*}} & \str{abab} & T & \regex{a{*}} & \str{aab} & F \\
         \regex{(ab){*}} & \str{aabab} & F & \regex{a{*}} & \str{aaa} & T \\
         \bottomrule
    \end{tabular}
    }
    \caption{
    We test a model’s ability to learn an instruction language (here, of RegExs), by training on examples of instruction + data pairs, then testing on novel instructions. Each RegEx can be seen as an instruction for a different matching task.
    Note that \emph{no} examples of the test RegExs are seen in training; rather the model must interpret the RegEx instructions themselves to understand the test tasks.
    \label{intro}
  }
\end{table}

Recent years have seen an increased interest in instruction learning~\citep{Weller2020LearningFT}
where a model learns to perform unseen tasks at test time in a zero-shot manner
given only a prompt containing instructions. This style of learning is an important feature of flexible and general intelligent systems.

Instruction learning stands in contrast to the traditional machine learning paradigm of \emph{example learning}. In example learning, the model has access to input-output pairs during training time. For instance, in sentiment analysis with the goal to classify product reviews, the model has access to many examples of labeled reviews from which to learn the task. 
In instruction learning, a model that has never seen labeled sentiment analysis
data must perform sentiment analysis given only the explicit instruction ``Tell me whether this review is positive". 
In other words, the model learns to interpret the instruction language (here
English) in order to execute an instruction for a task it has never seen.


Most recent work has on instruction learning has been conducted in the context
of natural language instructions~\citep[\eg][]{Wei2021FinetunedLM,mishra2021crosstask,Sanh2021MultitaskPT,Zhong2021AdaptingLM}.
In this context, the complexity of natural language makes it difficult to draw \addedB{clear conclusions} about the kinds of instructions transformers can \addedB{learn to} interpret. 
To remedy this, we adopt a synthetic approach by building an instructional environment based on interpreting regular expressions (RegExs) and use well-studied properties of RegExs to characterize transformer instruction learning capabilities. 

A RegEx is a specification of a formal language, \ie a set of strings of symbols. To avoid the confusion between an instructional language such as English and a formal language specified by a RegEx, we refer to the latter as an \emph{\rlang} (for regular language). In our work, we view a RegEx as an instruction for the task of deciding whether strings belong to the \rlang\ of the RegEx. 
We choose to study RegExs because they are well known, unambiguous (the \rlang\ decision problem is binary and always well-defined), and easy to compute (there exist linear-time algorithms to recognize whether a string belongs to a regex),
while also incorporating fundamental operations including iteration, disjunction, conjunction, and nesting.
This environment allows us to study instruction learning phenomena more precisely.

In our approach, we construct a collection of RegEx datasets (RegEx + string$\rightarrow$T/F) each for a specific RegEx instruction. 
We fine-tune a large T5-based text2text model on our datasets and find that the model is unable to correctly interpret many instructions in the test set.
Inspecting the \rlangs, RegExs, and string instances, we identify a number of properties that predict which RegEx instructions the model struggles with. 
Selecting RegExs with these properties, we construct a hard variant of the dataset which we call Hard RegSet. 
Our fine-tuned model achieves good performance on only 65.6\% of the test RegExs in Hard RegSet, leaving room for improvement.

\addedB{We find that} \added{instruction learning} models struggle with non-starfree \rlangs. This provides evidence that even large Transformers struggle with periodic \rlangs, a theoretical limit of transformers' attention mechanism~\citep{hahn-2020-theoretical} that \citet{Bhattamishra2020abilityOfSANetworks} further study in smaller models, in the example learning setting.

Our findings \addedB{in \S\ref{sec:attr_analysis} suggest four general implications that we expect will extend beyond the synthetic RegEx environment. First, instruction learning is harder when the underlying \emph{tasks} require modular counting (\eg keeping track of whether a quantity is even or odd). Second, it's harder when the \emph{instructions} themselves are not very precise, that is, they can be executed correctly in several different ways, forcing the execution engine to make and track choices. Third and perhaps least surprising, instructions involving executing and composing many individual operations or steps are harder to learn. Lastly, instruction learning is harder when correctly executing instructions requires keeping in memory a larger context of the partial execution thus far and making choices dependent on this longer history.}

In summary, our main contributions are:
\begin{ite}
\item We build the first (to our knowledge) fully synthetic environment for systematically investigating instruction learning.
\item We identify \addedB{general} properties that make instruction learning hard for a large T5-based model \addedB{and suggest broader implications beyond the RegEx environment.}
\item We show that previously \addedB{noted} limitations of \addedB{small} transformer models also apply to large models \eat{\addedB{and even}} in the instructional setting.
\item We construct a challenging dataset \addedB{(RegSet)} based on our findings to serve as a benchmark for future instruction-following models. 
\end{ite}

\section{Related work}

\subsection{Learning instructions} 

While large language models such as GPT3~\citep{Brown2020LanguageMA} have shown some intrinsic ability to understand task instructions expressed in natural language, those capabilities have been found to be rather limited~\citep{Efrat2020TheTT}. 
This has led researchers to ask whether models can be {\it trained} to understand instructions more reliably, creating the sub-field of instruction learning.
Our formulation and approach aligns closely with \citet{Weller2020LearningFT} who build \textsc{Zest}, a benchmark for natural language instruction following. 
In their work, they fine-tune language models on natural language tasks paired with instructions and test on new tasks. 
They find that T5, a large pre-trained text2text model~\citep{raffel2020exploring}, does poorly on their benchmark. 
Our work is distinct from this study in that we adopt a highly-controlled synthetic environment that allows us to perform an in-depth analysis of formal instruction properties. 

Following this work, numerous studies~\citep{mishra2021crosstask, Zhong2021AdaptingLM, Wei2021FinetunedLM, Sanh2021MultitaskPT} have shown improvement in the natural language instruction learning setting by fine-tuning a large pre-trained language model on collections of tasks, each annotated with a natural language instruction. Though our work differs in domain and goals, we adopt the framework of fine-tuning on instruction-annotated datasets to train an instruction learner.

\eat{
Because the common framework for training an instruction learner requires the model to generalize to new tasks with new instructions, our work draws on the rich body of literature on compositional generalization. 
Work on the SCAN dataset~\citep{Lake2018GeneralizationWS}, for instance, revealed major shortcomings in neural sequence models' systematic generalization ability. 
SCAN and subsequent work~\citep[\eg][]{Keysers2020MeasuringCG, shaw-etal-2021-compositional} has focused on the semantic parsing task, mapping an input (e.g., ``jump twice'') to an output (e.g., ``\texttt{JUMP JUMP}'') with a 1-to-1 mapping $x \rightarrow y$, 
\ie learning a single function from examples.
In contrast, instruction learning learns a functional {\it language}
such that arbitrary functions $f(x) \rightarrow y$ can be understood
from a description alone.
In this sense, RegSet generalizes the SCAN task. 
}

RegExs can be viewed as simple programs to execute on data (here strings), and
the task can thus be viewed as predicting the results of code execution.
Earlier work has shown that transformers can emulate (the results of) 
deductive reasoning over a set of natural language-like rules 
with limited expressivity~\citep{Clark2020TransformersAS}. 
Similarly, RegEx instructions include compositions of simple operations such as disjunction and concatenation as well as more complex rules such as iteration (the Kleene star ``*'' operator) and nesting (``()'').  

\eat{
In the program synthesis domain,~\citep{Nye2021RepresentingPP} combines search
for a program  with neural execution of a partial program (a lambda expression).
However this executor was a modular system custom-designed for their syntax,
and used to guide search rather than explicitly study a transformer's learning
abilities.}

\subsection{Compositional generalization}

Because the common framework for training an instruction learner requires the model to generalize to new tasks with new instructions, our work draws on the rich body of literature on compositional generalization. 
The SCAN dataset \citet{Lake2018GeneralizationWS}, for instance, revealed major shortcomings in neural sequence models' systematic generalization ability. Subsequent work has yielded a number of studies that attempt to identify properties of datasets~\citep{Keysers2020MeasuringCG, shaw-etal-2021-compositional} and instances~\citep{Bogin2022UnobservedLS,tamari2021dyna} that make generalization hard
and use these properties to construct hard generalization splits. 
The majority of these works focus on the domain of semantic parsing. 
Our work builds upon this literature by introducing the instructional setting for studying compositional generalization.
In addition, we use our setting to test the hypothesis put forward by \citet{Bogin2022UnobservedLS} that unseen local structures make generalization harder.

In {an approach close to ours}, \citet{Richardson2021PushingTL} use well-studied computational problems (SAT) to sample hard instances of deductive reasoning. Our work shares the goal of using formal properties to generate hard examples. Their work, however, is oriented towards demonstrating that random samples are often trivial, whereas we are interested in testing a number of hypotheses on what makes examples hard. 

\subsection{RegEx expressions and transformers}

A number of studies have used formal language theory to investigate the expressive and computational power of transformers and other models \added{(for a review, see \citet{merrill2021formal})}. This includes theoretical work~\citep{hahn-2020-theoretical} and empirical studies~\citep{Bhattamishra2020abilityOfSANetworks}. Many of these studies \added{focus on specific \rlangs and other well-studied classes of formal languages} and specifically give evidence that certain types of \rlangs are hard for transformers \added{to learn. The theoretical studies have often assumed simplifications (e.g., hard-attention, smooth activation functions). The empirical studies tend to use small, toy-sized transformers (e.g., \citeauthor{Bhattamishra2020abilityOfSANetworks} use 1-4 layer transformers with 4 heads and dimension up to 32).} Our work investigates findings from these empirical studies under a setting where instead of learning a single \rlang with a model, we task the model with learning how to interpret \rlangs in general. Additionally, we use a \added{commonly used} large transformer model (based on T5-Large) with much higher capacity than the ones used in these studies. 

Additionally, \citet{Dan2022UnderstandingRG} use also use \rlangs to study generalization by showing that RNNs struggle to generalize under distributional shifts when recognizing \rlangs, then using an auxiliary task (modeling a DFA) to improve generalization. 

\section{RexEx instruction learning task}

In this section we give a brief description of regular languages and RegExs, and formulate the general problem of instruction learning.

\subsection{Regular languages and expressions}

\added{We use standard terminology from the formal languages literature~\cite[e.g.][]{harrison1978introduction}, focusing on regular languages and regular expressions. For completeness, formal definitions and properties are included in Appendix~\ref{app:rlangs}. We briefly describe the key concepts useful for understanding our results.}

\added{We work with strings over the alphabet $\{a,b\}$. A set of such strings is called an \rlang\footnote{We use \rlang as a short form of regular language, in part to distinguish it from another language that plays a key role in this study, namely the language of instructions.} if it's the empty set; a singleton set containing $a$, $b$, or the empty string $\varepsilon$; the union of two \rlangs; the element-wise concatenation of two \rlangs; or the \emph{Kleene star} of an \rlang, which is defined as zero or more occurrences of elements from that \rlang.}

\added{A regular \emph{expression} (RegEx) is an often succinct and non-unique specification of a regular language. Its string nature makes it suitable for text2text models. We denote the language specified or ``expressed'' by a RegEx $r$ as $L_r$. The RegEx $a$ represents the singleton \rlang $\{a\}$ (similarly for $b$ and $\varepsilon$), $r|s$ represents the union of $L_r$ and $L_s$, $rs$ represents the element-wise concatenation of $L_r$ and $L_s$, and $r\regex{*}$ represents the Kleene star of $L_r$. Parentheses are used to indicate the order of operations, \textit{e.g.} in \regex{(a|bab)b{*}}. We refer to union, concatenation, and Kleene star as \emph{compositional operators}.}

\subsection{Instruction learning}
\label{sec:instruction_learning}

\newcommand{\TTrain}{T^{\mathrm{train}}}
\newcommand{\TTest}{T^{\mathrm{test}}}

We start by formally describing what we mean by \emph{instruction learning}, as opposed to the standard ML paradigm of learning a task from its input-output examples, which one might refer to as \emph{learning from examples}. One can view instruction learning as the ML approach to the overall task of \emph{instruction following}, which seeks to be able to interpret instructions expressed in an instruction language in order to solve novel tasks.

More concretely, in both instruction following and instruction learning, one begins with an instruction language $\instrlang$ that can be used to describe tasks, and a set of tasks $T = \{t_1, \dots, t_m\}$. Each task $t_j \in T$ is paired with a descriptive instruction $\instr_j \in \instrlang$ as well as with input-output examples $D_j = \{(x_{ji},y_{ji}) \mid 1 \leq i \leq N_j\}$. 

The goal of \emph{instruction following} is to be able to solve \emph{unseen} tasks given only their description $\instr_j \in \instrlang$, i.e., to be able to map the description $\instr_j$ of a novel task and an input $x_{ji}$ for it to the correct output $y_{ji}$. In \emph{instruction learning}, one aims to achieve this in a data driven way, by training a model. A key aspect that makes this feasible is compositionality in $\instrlang$, i.e., pieces of $\instrlang$ can be learned at training time and combined in different ways at test time. The training data for instruction learning consists of a subset $\TTrain \subsetneq T$ of the tasks, where each task $t_j \in \TTrain$ is specified via its descriptive instruction $\instr_j \in \instrlang$ and input-output examples $D_j$. At test time, one is given the description and an input for a task in $\TTest = T \setminus \TTrain$.

In this notation, standard single-task example learning corresponds to the (degenerate) case where $\TTrain = T = \{t_1\}$, $\TTest = \TTrain$, and $\instr_1 = \epsilon$ is the empty string denoting the \emph{null} instruction. In other words, the model has no descriptive instruction available to help learn $t_1$; it must be learned solely from input-output examples.\footnote{In standard learning, including an identical non-null instruction in every train and test example is not helpful as it does not form a discriminative feature.} Similarly, standard multi-task learning corresponds to $T$ having multiple tasks $t_j$ and the corresponding instructions $r_j$ may be either null or a short prefix identifying the task (e.g., as proposed by \citet{raffel2020exploring}). However, all tasks queried at test time are already seen and learnt during training time ($\TTest = \TTrain$). Lastly, \emph{prompt based zero-shot models} can be viewed as a (degenerate) case at the other extreme, where each $t_j$ has the associated discrete or continuous prompt as the instruction $\instr_j$, but no input-output examples, i.e., $D_j = \phi$. In this case, it is critical for the model to already understand the instruction (or prompt) language $\instrlang$. In practice, one uses a large language model as the model, and the natural language it is trained on (e.g., English) or its continuous embedding as $\instrlang$.

Applying this formalism to the case of our RegEx instruction learning challenge, we view $\instrlang$ as the language consisting of all RegExs, i.e., an instruction is a RegEx. Each task $t_j$ is associated with a specific \rlang $L_j$ and involves deciding whether an input string belongs to $L_j$. The instruction $r_j$ associated with $t_j$ is a RegEx\footnote{Note that our instructions are not written in natural language. One can, in principle, convert each RegEx instruction into an equivalent, even if cumbersome, English description.} describing $L_j$. The input-output examples consist of strings paired with 1 or 0, depending on whether or not the string belongs to $L_j$.

We should note that this is different from prior work on RegEx learning~\citep{Bhattamishra2020abilityOfSANetworks}, which has focused on learning a model for a specific RegEx (e.g., well-studied regular language classes such as parity languages, which can be characterized as a single RegEx). In our notation, this corresponds to the single-task setup mentioned earlier, where $\TTrain = \TTest = \{t_1\}$ and $\instr_1 = \epsilon$. Similarly, works such as SCAN~\citep{Lake2018GeneralizationWS} can also be viewed naturally as one of two (degenerate) extreme cases of instruction learning, as follows. Let $z_1, z_2, \ldots$ denote SCAN inputs such as ``jump twice''. We can view SCAN as involving only one meta-task ($T = \{t_1\}$) with an (implicit) instruction $\instr_1$ conveying something like ``convert the input to a sequence of executable steps'', and with task inputs $x_{1i} = z_i$. Alternatively, one can view SCAN as having as many distinct tasks $T = \{t_1, t_2, \ldots\}$ as SCAN inputs, the associated instructions being $\instr_j = z_j$, and the task inputs $x_{ji}$ being null (i.e., $D_j = \phi$) because, in this view, $\instr_j$ fully determines the output $y_{ji}$ (\ie instruction $\instr_j$ can be executed in only one way, leading to the output $y_{ji}$).

\section{RegEx attributes}

In this section we define a number of attributes of instances of our RegEx instruction learning task. In later sections we use these to characterize our datasets and identify what makes our task difficult.

\subsection{Language-level attributes}

We use the term language-level attributes to refer to properties of \rlangs that are invariant to how the \rlang is expressed. For instance, \regex{a{*}} and \regex{aa{*}|(aa){*}} express the same \rlang, and thus ``presence of a union operator'' is not a language-level attribute. On the other hand, the number of strings of length $< 15$ in the \rlang remains the same no matter how the language is expressed as a RegEx and is therefore a language-level attribute.

\paragraph{Starfreeness.}
A well-known complexity measure of an \rlang is whether it is \emph{starfree} \added{\citep{mcnaughton1971counter}}.
An \rlang is starfree if there exists a RegEx for the \rlang whose operators include only 
union, concatenation, and set complement (notably, these operators exclude the Kleene star).
For instance, \regex{a{*}} can also be expressed as $(\emptyset^c\mathrm{b}\emptyset^c)^c$,\added{\footnote{The \rlang corresponding to $r^c$ is the set complement of $L_r$, the \rlang of $r$.}}
so the \rlang expressed by \regex{a{*}} is starfree.
Previous work~\citep{Bhattamishra2020abilityOfSANetworks} has shown that small transformer models struggle to generalize when trained on a non-starfree \rlang 
(whereas LSTMs are able to generalize perfectly). 

\paragraph{Size.}
Since an \rlang is a set of strings, one attribute of interest is the size of the set.
For many \rlangs, this size is infinite. For practicality and to distinguish between \rlangs of infinite size, we define the \emph{size} of an \rlang to be the number of strings in the set with length up to 14. 
By this definition, the maximum size of an \rlang in our dataset is $2^{14}=16384$.

\subsection{Expression-level attributes}

Expression-level attributes are specific to a RegEx, but not the \rlang they express. For instance \regex{a{*}} and \regex{aa{*}|(aa){*}} express the same \rlang, but only the latter uses a union operator. Thus having a union operator is an expression-level attribute.

\begin{figure}[t]
\centering
\begin{forest}
  [$*$, tikz={\node [draw,black,fit to=tree] {};}
    [$\cup$, s sep=12mm, tikz={\node [draw,black,fit to=tree] {};} 
        [\regex{a}]
        [$\cdot$, tikz={\node [draw,black,fit to=tree] {};} 
            [\regex{b}, tikz={\node [draw,black,fit to=tree] {};} ]
            [$\cdot$ [\regex{b}][\regex{a}]]]]]
\end{forest}
\caption{Regex \regex{(a|bba){*}} contains sub-expression \regex{a|bba} which contains sub-expression \regex{bba} which contains sub-expression \regex{b}.}
\label{fig:subexp}
\end{figure}
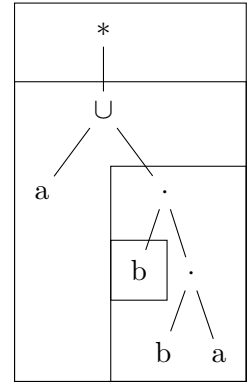

\paragraph{Composition.}
RegExs are constructed by recursively composing RegExs together using union, concatenation, and star operators. To study the effect of the amount of composition in an RegEx on its difficulty, we define the amount of \emph{composition} in an expression as the number of operators used to construct it, \textit{e.g.} the expression \regex{(a|bba){*}} has 3 compositions: 1 star, 1 union, and 2 concatenations. We conjecture that more composed RegExs are more difficult due to their increased complexity.
    
\paragraph{Unseen sub-expressions.}
We refer to a RegEx that is an argument to a RegEx operator as a \emph{sub-expression}, \eg \regex{a|b} is a sub-expression of \regex{(a|b){*}}. 
Prior work~\citep{Lake2018GeneralizationWS, Keysers2020MeasuringCG, Bogin2022UnobservedLS} has shown that models often fail to generalize to new compositions of atomic components, even when all the components have been seen during training time.
To investigate this in our own datasets, we define the \emph{unseen sub-expressions} of a RegEx w.r.t. a dataset
as the set of sub-expressions contained in the RegEx that are not contained in any of the RegExs in the dataset. In cases where a training set contains all atomic symbols, this becomes a measure of unseen compounds in a test set RegEx. 

\subsection{Instance-level attributes}

Instance-level attributes are attributes that depend not only on the RegEx or language expressed by the RegEx, but also on the string being recognized.

\paragraph{Execution states (ES).} 
An \rlang can be equivalently defined as the strings accepted by a deterministic finite automaton (DFA).
We define \emph{execution states} (ES) of a string $x$ with respect to a RegEx $r$ to be the number of unique states in $r$'s \rlang $L_r$'s minimal DFA that are visited while recognizing $x$. This is closely related to the notion of \emph{state-complexity}~\citep{yu2001state}. 
Since each \rlang has a unique minimal DFA, this property is invariant with respect to the RegEx used to express the \rlang. 
We choose this metric as a way to measure how much space is required to execute the string recognition problem, with the intuition that keeping track of more states is harder.

\paragraph{Ambiguity.} We define a similar metric, \emph{ambiguity} as the maximum number of sub-expressions any token refers to as a string is processed in either direction. For instance, given RegEx $r = \regex{a|ab|abb}$ and string $x = \str{abb}$, processing left to right, $\str{a\ldots}$ could refer to any of the three disjoint sub-expressions, \str{ab\ldots} could refer to either of the last two sub-expressions (\regex{ab|abb}), and finally, \str{abb\ldots} could refer only to the last sub-expression (\regex{abb}). The same can be done considering $x$'s elements in reverse order. We take the minimum value for the two directions to be the \emph{ambiguity} of $x$ w.r.t.\ $r$. This metric is intended as an expression-level analogue of ES, as we measure the computational space complexity of the execution, only here we do not allow the implicit conversion to a minimal DFA. 

\section{RegEx datasets}

Our datasets consist of collections of triples according to the formulation given in \S\ref{sec:instruction_learning}, where each triple $(r,x,\ell)$ contains a regex $r$, a string $x \in \{\str{a},\str{b}\}^*$, and a label $\ell \in \{0,1\}$ indicating whether $x \in L_r$ where $L_r$ denotes the \rlang specified by the regex $r$. Some examples are given in Table~\ref{tab:dataset}. The RegEx instruction learning task is to learn to predict $\ell$ given $(r,x)$ as input. 

\begin{table}[t]
  \centering
  \begin{tabular}{lll}
    \toprule
    Regex $r$ & String $x$ & $x \in r$? \\
    \midrule
    \regex{aba|b{*}} & \str{aba} & True \\
    \regex{(b|a){*}} & \str{abbaba} & True \\
    \regex{bab|b}    & \str{ab} & False \\
    \bottomrule
  \end{tabular}
  \caption{Input to the model is ``$(x,r)$'', output from the model is ``True''
  or ``False''.}
  \label{tab:dataset}
\end{table}

\subsection{Sampling RegExs and strings}

We sample RegExs $r$ such that all corresponding \rlangs $L_r$ are distinct within the dataset. 
Furthermore, $r$ is chosen such that it uses the minimum number of compositional operators needed for representing $L_r$. 
We achieve this by generating RegExs from small to large, starting with 0-operator RegExs before moving to 1-operator RegExs and so forth. We keeping track of what \rlangs have been generated so far as we go. 
Formally, let $\mathcal L_n$ be the set of \rlangs expressible with $n$ operators but not expressible with less than $n$ operators. For each $n$, we find all $L\in\mathcal L_n$ and randomly choose an RegEx $r$ with $n$ operators that expresses $L$. By constraining the RegExs to express unique \rlangs, we avoid the problem of over-sampling certain \rlangs: A naive sample would produce overwhelming numbers of redundant RegExs that express the same \rlang.  

We limit the maximum number of compositional operators in each $r$ to $6$,
choosing RegExs approximately uniformly at random with respect to the number of operators. Specifically, we separate all possible RegExs into bins based on the number of operators they have. From each bin, we randomly sample the same number of RegExs, with the exception of small bins that contain too few RegExs, in which case we sample as many as we can. The algorithm for our sampling strategy is given in \S\ref{sec:sample}.

For strings $x$, we limit the maximum length to $15$ and sample each length approximately uniformly at random (again, since there are fewer strings with shorter length, we sample fewer of those).
For each $r$, we sample both strings that match $r$ and strings that do not. 

\subsection{Exploration RegSet}
\label{subsec:exploration_regset}

To investigate which attributes make RegEx instruction learning hard, we construct an ``exploration'' training set and a corresponding large test set with which to explore what our model has learned. We refer to this as the Exploration RegSet.

Our training set contains 1000 RegExs, each with 20 strings. 
We choose our RegEx-to-string ratio to be the ratio that results in the best model given a budget of 20K training examples. 
Because the number of strings in (or not in) an \rlang varies and sometimes is less than 10, 
it is not possible to fully balance the data set. 
However, we choose the maximum number of strings from the minority class up to 10 so that the dataset is as balanced as possible while maintaining 20 strings per RegEx.
We also set aside 200 validation RegExs each with a positive and negative string for model selection purposes.

Our test set contains 500 RegExs, each expressing an \rlang not found in the training set.
Each regex in the test set is paired with a balanced set of positive and negative strings up to length 15. We are able to balance this set by relaxing the constraint on the number of strings per RegEx, instead sampling $\min(|L_r|, |L_r^c|, 1000)$ strings per regex.
We do this to achieve a balanced approximation of an exhaustive set of strings to test. On average, each RegEx includes 200 strings.

We use the properties that make instruction learning hard in Exploration RegSet to create a hard version the dataset which we call \textbf{Hard RegSet}. Details of Hard RegSet are given in \S\ref{sec:hard_regset}. 

\begin{table}[t]
    \centering
    \begin{tabular}{lrr}
    \toprule
    Attribute & Exploration & Hard \\
    \midrule
    Regexs (\#) & 1,000 & 1,000 \\
    Instances/RegEx (mean) & 20 & 20 \\
    Starfree (\%) & 86.1 & 0 \\
    R-lang. size (med) & 8 & 368 \\
    Compositions (mean) & 4.9 & 6.0 \\
    ES (mean) & 3.4 & 5.4 \\
    String length (mean) & 7.6 & 10.5 \\
    \bottomrule
    \end{tabular}
    \caption{Comparison of attribute statistics for Exploration and Hard training sets. \added{The number of unseen sub-expressions is omitted as it is defined for a test set w.r.t.\ a training set.}}
    \label{tab:my_label}
\end{table}

\section{Experimental setup}
\label{sec:setup}

We describe the datasets, model, and performance metrics used in our experimentation.

\subsection{Datasets} 

We will use Exploration RegSet (\S\ref{subsec:exploration_regset}) to explore which properties make RegEx instruction learning hard (\S\ref{sec:attr_analysis}). We will then use these properties to construct the more challenging Hard RegSet (\S\ref{sec:hard_regset}), and demonstrate that the model struggles on it in both in-domain and out-of-domain settings.

\subsection{Model} 
We choose ByT5-Large\footnote{\addedB{In our explorations} with in-context learning using GPT-3 \citep{Brown2020LanguageMA}, \addedB{the model does no better} than random guessing.} (1.2B parameters)~\citep{Xue2022ByT5TA} as our main model. ByT5 is a T5-based pre-trained transformer model with character-level tokenization---which is helpful in avoiding issues with tokenizing synthetic strings like ``aabba|bb(a){*}".
We choose a model pre-trained on natural language (as opposed to a randomly initialized transformer) because, though our RegEx task has little resemblance to natural language, previous work has suggested that pre-training imbues performance benefits even for unrelated tasks~\citep{Krishna2021DoesPF, Maennel2020WhatDN}. 

Using ByT5 as a text-to-text generation model, we feed $r\ x$ to the model as a string, separating $r$ and $x$ with a space, and train the model to output the strings ``True" or ``False". We train on the full train sets for 200 epochs, using the validation accuracy to select the best model. We use a learning rate of $5\cdot10^{-5}$ and batch size 32.

\subsection{Metrics}
Let $D$ denote an evaluation set for the RegEx instruction learning task, and $M$ a model for it. We define $M$'s accuracy on an instance $(r,x) \in D$, denoted $\acc_M(r,x)$, as $1$ if $M$ correctly predicts whether $x$ matches $r$, and $0$ otherwise. $M$'s accuracy for a RegEx $r$ is $\acc_M(r) = \avg_{x : (r,x) \in D} \acc_M(r,x)$, where $\avg$ denotes arithmetic average.

We are interested in measuring how well $M$ learns to interpret each RegEx. To this end, we use metrics that operate at the level of RegExs. Due to the synthetic---and thus noise-free---nature of our datasets, we expect $M$ to \emph{perfectly} learn every expression given sufficient training data, at least in the i.i.d.\ setting and also in reasonable generalization settings. In other words, the upper bound for $\acc_M(r)$ is 100\% for each $r$. To measure how well $M$ does relative to this upper bound, we define two metrics, both of which give equal weight to all RegExs in $D$, regardless of how many test strings each has.

Our main metric is \textbf{Mean RegEx Performance at k}, defined as the fraction of RegExs on which $M$'s accuracy is at least $k$ (treated as a percentage), i.e., $M$'s performance if the goal is to learn each RegEx to an accuracy of at least $k$:
\begin{equation}
    \perf_M@k = \avg_{r \in D} \, I \big(\acc_M(r) \geq k \big)
\end{equation}
where $I(\cdot)$ denotes the indicator function\footnote{$I(c)$ is $1$ if the condition $c$ is satisfied and $0$ otherwise.} and, with slight abuse of notation, we use $\{r \in D\}$ as a shorthand for $\{r \mid (r,x) \in D\}$. We will drop the subscript $M$ from $\perf_M$ when the model is clear from the context. $\perf@100$ thus refers to the fraction of RegExs $r$ that are learned \emph{perfectly} (as assessed by all strings tested for $r$ in $D$). As noted above, the upper bound for $\perf@100$ is 100\% given sufficient training data, by virtue of the noise-free nature of our datasets. Since this metric is somewhat \added{strict} from a machine learning perspective, we use {${\perfbf\mathbf{@90}}$} as our main metric, and also track a more lenient metric, $\perf@80$.

We also report a secondary metric referred to as \textbf{Mean RegEx Accuracy}, defined as $M$'s accuracy on a RegEx $r$, averaged across all $r$ present in $D$:
\begin{equation}
    \acc_M = \avg_{r \in D} \, \acc_M(r)\,.
\end{equation}
As before, we drop the subscript $M$ from $\acc_M$ when the model is clear from the context. Note that this metric does not distinguish between two RegExs that are learned to accuracies of 90\% and 50\%, respectively, vs.\ two RegExs that are each learned to an accuracy of 70\%. Further, when the number of string per RegEx is identical across the evaluation set, $\acc_M$ simplifies to the standard instance-level accuracy of the dataset rather than a RegEx-level metric. For these reasons, this metric is less desirable, but we include it for completeness.

\section{Results: Which attributes make instruction learning hard?}
\label{sec:attr_analysis}

\begin{table}[t]
    \centering
    \begin{tabular}{lcc}
    \toprule
    Attribute classes & Class 1 & Class 2 \\
    \midrule
    Starfree/non-starfree &	91.9 & \bf 71.9 \\
    Small/big \rlang & 95.1 & \bf 57.5 \\
    Low/high composition & 95.2 & 80.3 \\
    None/has unseen exprs & 95.9 & 87.5 \\
    \bottomrule
    \end{tabular}
    \caption{Summary of $\perf@90$ results for language-level {(top 2)} and expression-level {(bottom 2)} attributes. Language-level attributes we measure have a larger impact on performance than expression-level attributes.}
    \label{tab:results_summary}
\end{table}

We first train our model on the Exploration Train set and evaluate on the Exploration Test set. 
The model achieves a respectable mean RegEx accuracy of $\acc_M=97.1\%$. 
The number of RegExs learned to at least 90\% accuracy, however, is a more modest $\perf@90$ of $89.6\%$, meaning that the model hasn't quite learned the how to interpret RegExs. 
By design, the Exploration test set's large size (500 RegExs with $200$ strings each, on average) allows us to analyze the model's errors in detail and assess which attributes contribute to the hardness of instruction learning. 
Table~\ref{tab:results_summary} summarizes $\perf@90$ scores for language- and expression-level attributes. We discuss the details of our findings below.

\begin{figure}[t]
  \centering
  \includegraphics{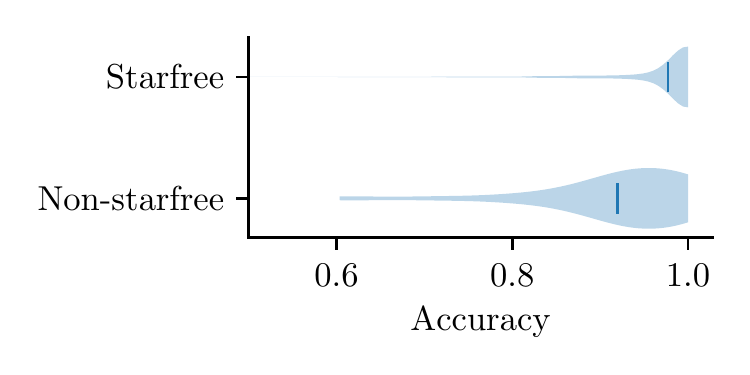}
  \caption{Non-starfree \rlangs are
  hard for our model. We compare mean RegEx accuracy between starfree and non-starfree \rlang in our test set using a violin plot. The blue region represents the distribution of the data, with a blue bar showing the mean. We find that, while starfree \rlangs have $\perf@90=91.9\%$, non-starfree \rlangs are learned at a significantly lower rate of $\perf@90=71.93$. } 
  \label{fig:sf}
\end{figure}  

\paragraph{Non-starfree \rlangs are hard.}
Among the \rlangs in our Exploration training set, 86.1\% are starfree.
We find that non-starfree expressions are significantly harder to interpret for our model, as shown in Figure~\ref{fig:sf}.
We conjecture that these RegExs are harder for our model for the same reason that \citet{Bhattamishra2020abilityOfSANetworks} find that small transformers fail to learn and generalize even the simplest non-starfree \rlangs such as $(aa){*}$. This finding also aligns with the theoretical prediction that transformers struggle to model periodicity---a common feature among non-starfree \rlangs~\citep{hahn-2020-theoretical}.
As our own addition to this body of work, our results show that, despite increased capacity, large models \added{(as used in practice, without simplifying assumptions)} still struggle with non-starfree \rlangs under the instructional setting.
Our results suggest that even large transformers may struggle broadly with instructions that require modeling periodicity and modular counting, such as keeping track of whether a quantity is even or odd. 


\begin{figure}[t]
    \centering
    \includegraphics{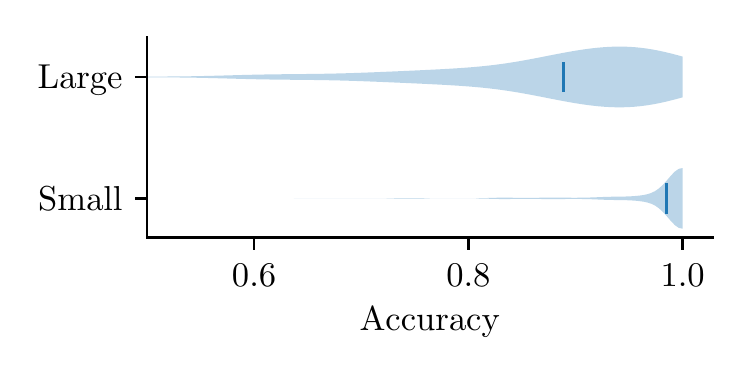}
    \caption{The size of an \rlang is highly predictive of difficulty. We show the difference in accuracy between large \rlangs with more than 64 strings, and small \rlangs with at most 64 strings. The drop in learned RegExs is dramatic with $\perf@90=95.1$ for small \rlangs but only $\perf@90=57.5$ for large \rlangs.}
    \label{fig:size}
\end{figure}

\paragraph{Big regular languages are hard.}
\label{sec:balance}
Because we limit the amount of composition in our RegExs and do not use the complement operator, the \rlangs in our dataset tend to have small sizes.
As a result, the median \rlang size in the Exploration training set is 8. 
As shown in Figure~\ref{fig:size}, we find that our model tends to struggle with \rlangs with more strings. 
We speculate that balanced \rlangs become computationally non-trivial for transformers to decide. Similarly to how \citet{Richardson2021PushingTL} find that SAT problems become non-trivial for models when the number of variables to clauses is such that the ratio of satisfiable to non-satisfiable problems is balanced (i.e., on problems within the critical phase-change regions studied in random $k$SAT~\citep{selman1996generating}), we propose that larger \rlangs (at least ones with size near $2^{13}$, half the strings with length less than 15) are harder to approximate by observing trivial patterns. 

Speculating further, we hypothesize that models may struggle generally with less precise instructions. Small \rlangs have a very narrow scope of interpretation: only a few strings match the specification. Since these are the easiest for the model, we postulate that more generally, precise instructions with few possible interpretations are easier for models and conversely, more general or abstract instructions are more challenging.

\begin{figure}[t]
  \centering
  \includegraphics{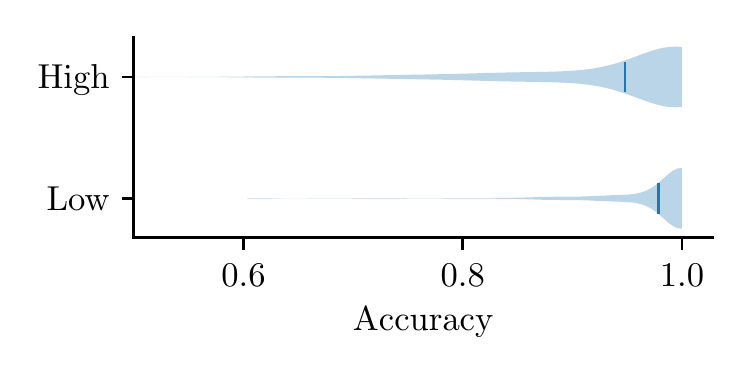}
  \caption{More compositions make expressions somewhat harder. We compare mean accuracy of RegExs with 5 or fewer operators (low-composition) and mean accuracy for RegExs with 6 operators (high composition). The drop in $\perf@90$ is modest: from 92.9\% for low-composition RegExs to 80.3\% for high-composition RegExs.}
  \label{fig:comp}
\end{figure}

\paragraph{More composition makes expressions harder.}
The Exploration training set contains 55 RegExs with fewer than 4 compositions and 315 RegExs for each depth from 4 to 6. 
The fewer numbers of shallower RegExs is due to the fact that the number of expressions increases exponentially with number of compositions. 
As seen in Figure~\ref{fig:comp}, we find that accuracy decreases somewhat for expressions with more composition.
This finding aligns well with intuition because more composed strings tend to be longer and the complexity of the RegEx string recognition algorithm scales linearly with input length~\citep{Thompson1968ProgrammingTR}. 

Speculating beyond the regex domain, we hypothesize that instructions composed of many sub-instructions are more challenging for models like T5 compared to less compositional instructions, but that other factors (like specificity or periodicity) likely play a larger role in determining difficulty.  
Though we do find that more composed expressions are harder for the model, the difference in accuracy is not great, so we ultimately do not select high composition as a criteria in building our Hard RegSet.

\begin{figure}[t]
  \centering
  \includegraphics{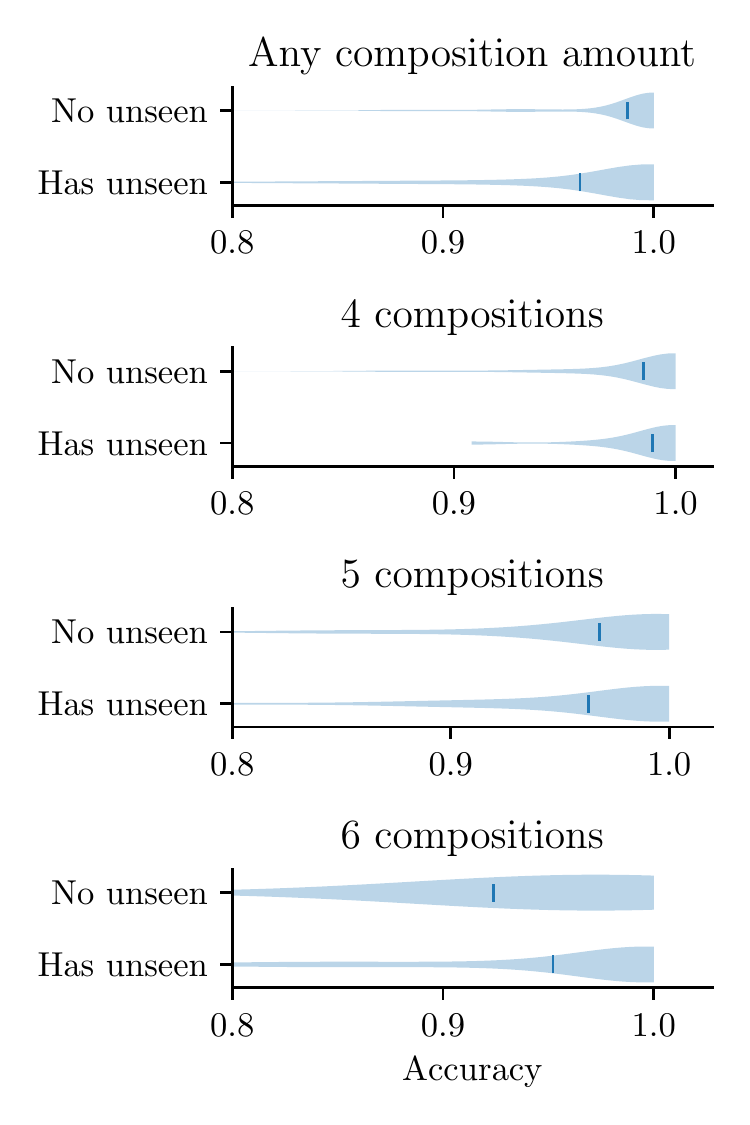}
  \caption{We compare mean accuracy between expressions with and without at least 1 unseen sub-structure. Without controlling for amount of composition (top violin plot), unseen sub-expressions appear to make expressions harder. However, controlling for amount of composition, (the bottom 3 plots) these effects disappear. The additional difficulty of more composed expressions accounts for the additional difficulty of expressions with unseen sub-expressions.}
\end{figure}

\paragraph{Unobserved local structures do not contribute significantly to hardness.} 
33.4\% of the expressions in the Exploration test set contain substructures not seen in the training set. 
\citet{Bogin2022UnobservedLS} propose that unseen local structures make compositional generalization hard. 
Inspired by this work, we hypothesize that the same effects occur in our setup as well.
We say that a RegEx contains unseen local structures if it has \emph{unseen sub-expressions} with respect to the training set. 
Upon first inspection, RegExs with unseen sub-expression appear harder, which would suggest support for our hypothesis. 
However, there is a confounding variable: deeper RegExs are more likely to have unseen substructures. 
When we control for the amount of composition in the RegEx, the observed effect disappears.\footnote{This is an example of Simpson's Paradox (\url{https://en.wikipedia.org/wiki/Simpsons_paradox}).}
We therefore conclude that the unseen local structures are not a major factor in what makes instruction learning hard in the RegEx domain. 
There are many possible explanations for this, including the possibility our model is able to generalize well in settings like ours with very few atomic symbols and operators.

\begin{figure}[t]
    \centering
    \includegraphics{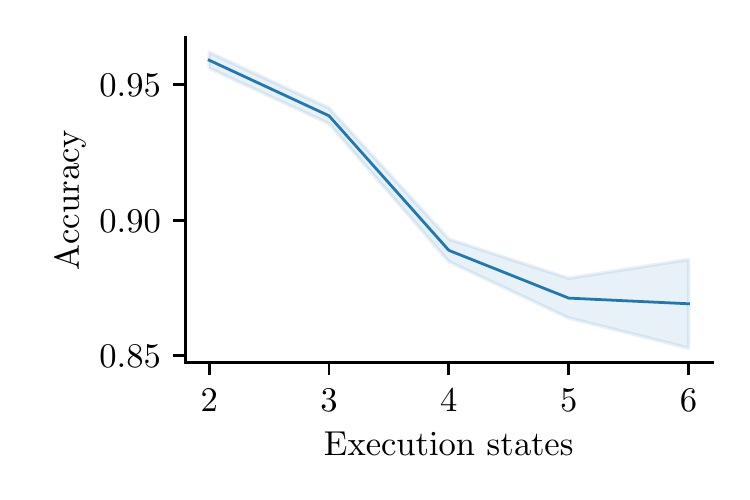}
    \caption{Accuracy decreases for high-ES strings. We group test set instances by ES and plot average accuracy (filled in area shows standard error), leaving off groups with fewer than 100 instances.}
    \label{fig:execution_states}
\end{figure}

\paragraph{Instances requiring many execution states are hard.}
We observe in Figure~\ref{fig:execution_states} that performance decreases for RegEx-string pairs where many distinct states are visited by the minimal DFA recognizing the string. 
Interestingly, \emph{ambiguity}, which can be viewed as an expression-level version of ES, does not produce a good predictor of hardness.

\added{Intuitively, being in many different states means that when processing the next character (say 0), one must act differently depending on how one arrived at that point, \ie depending on the prefix of the string up till that character. More states thus implies more of the prior context or history must be remembered and taken into account when processing the next character. Speculating beyond RegExs, we hypothesize that instruction learning is harder when determining the next valid step (while following the instruction) requires considering a longer context of prior steps.}

\begin{table}
  \centering
\begin{tabular}{lc}
\toprule
Regex&Accuracy (\%) \\
\midrule
\regex{b|(a|(a|b)b)^*   } & 34 \\
\regex{aa(a(a|b))^*|a   } & 44 \\
\regex{(b(b|ab))^*a^*   } & 50 \\
\regex{(a|bbb)^*b|a     } & 57 \\
\regex{(b(a^*aa|b))^*   } & 60 \\
\regex{((b|(a|b)a)b)^*|a} & 61 \\
\regex{b((b|a)a)^*a     } & 62 \\
\regex{b|(b(a|b))^*     } & 72 \\
\bottomrule
\end{tabular}

  \caption{The 8 lowest-scoring RegExs obtained by filtering the Exploration test set for large non-starfree \rlangs paired with strings with high ES.}
  \label{tab:interesting-hard}
\end{table}

\begin{table*}
  \centering
    \begin{tabular}{llc|cccc}
    \toprule
     & & & \multicolumn{4}{c}{\textit{Metric}} \\
    \textit{Training Set} & \textit{Evaluation Set} & & \hspace{2ex} $\acc$ \hspace{2ex} & $\perf@80$ & $\perfbf\bf@90$ & $\perf@100$ \\
    \midrule
    --          & Exploration & RND & 50.0 & \phantom{0}0.0 & \phantom{0}0.0 & \phantom{0}0.0 \\
    --          & Hard        & RND & 50.0 & \phantom{0}0.0 & \phantom{0}0.0 & \phantom{0}0.0 \\
    \midrule
    Exploration & Exploration & IID & 97.1 & 96.4 & \textbf{89.6} & 69.9 \\
    Hard        & Hard        & IID & 88.9 & 81.6 & \textbf{65.6} & 15.2 \\ 
    \midrule
    Exploration & Hard        & OOD & 77.2 & 52.8 & \textbf{23.4} & \phantom{0}2.0 \\
    Hard        & Exploration & OOD & 66.8 & 29.3 & \textbf{11.0} & \phantom{0}3.8 \\
    \bottomrule
    \end{tabular}
  \caption{Performance of the ByT5 model on Exploration and Hard RegSet datasets, in both in-distribution (IID) and out-of-distribution (OOD) settings. RND denotes the uniform random baseline. $\acc$ denotes Mean RegEx Accuracy (\%) and $\perf@k$ the percentage of RegExs with model accuracy at least $k$\% (\S\ref{sec:setup}). $\perfbf\bf@90$ is our main metric, under which the model struggles (65.6\%) on the IID Hard RegSet and performs very poorly (11.0\%-23.4\%) in the OOD settings.}
  \label{tab:model-performance}
\end{table*}

\section{Hard RegSet: a new challenge}
\label{sec:hard_regset}

Based on our findings for various attributes, we select the attributes that we believe contribute most to difficulty, namely starfreeness (or rather, non-starfreeness), \rlang size, and ES; in doing this, we follow the idea of \emph{salient variable} sampling from \citet{shin2019synthetic}. 
In selecting attributes we are careful to balance the trade-off between accuracy reduction and aggressiveness of the filter. 
For instance, although we do observe that high-composition RegExs tend to have lower accuracy, we do not choose this attribute because the effect is not especially large.  
Filtering the Exploration test set for non-starfree expressions of size greater than 64 and ES greater than 4 we achieve a reasonable sized set of 785 instances from 16 RegExs with $\acc{}_M = 72.2$. 
The 8 lowest-scoring RegExs from this group are shown in Table~\ref{tab:interesting-hard} for illustrative purposes. 
We therefore choose these settings to generate Hard RegSet:

\begin{quote}
Hard RegSet contains instances $(r,x,\ell)$ such that $r$ is not in the exploration training set, $r$ is non-starfree, the corresponding \rlang $L_r$ contains more than 64 strings, and the number of execution states (ES) for $x$ w.r.t.\ $r$ is greater than $4$.
\end{quote}

Hard RegSet is split into train, validation, and test sets\footnote{\addedB{Note that unlike some prior studies~\cite{ethics-dataset} that only consider a hard \emph{test} set, we also provide the corresponding hard \emph{training} and validation sets with identical distribution. This allows ruling out confounding factors such as distribution mismatch being the prime reason for the observed hardness.}} which match the sizes of the exploration train, validation, and test sets.

\subsection{Performance on RegSet}

Table~\ref{tab:model-performance} summarizes the performance of our model on both Exploration and Hard sets. Recall that our main metric in this instruction learning setup is $\perf@90$, \ie how many RegEx instructions does the model learn with an accuracy of at least 90\%. For completeness, we also include Mean RegEx accuracy (acc). The Random baseline, which predicts 0/1 with an equal probability, has an accuracy of 50\% and $\perf@90$ of zero.

We see that the model struggles on Hard RegSet even in the in-distribution setting (IID), achieving a $\perf@90$ score of only 65.5\%. As noted earlier, the upper bound is essentially 100\% due to the programmatic nature of the task. Closing the 34.5\% gap thus remains a challenge.

Further, in our out-of-distribution (OOD) settings, we train the model on the Exploration set and test on the Hard set, and vice versa. Here the model achieves $\perf@90$ scores of only 23.4\% and 11.0\%, respectively. Even the raw accuracy scores are quite low (77.2\% and 66.8\%) relative to the Random baseline. In other words, the model really struggles to generalize to OOD RegEx instructions, even though these instructions use the same primitives (the few basic RegEx operators) and have similar syntactic properties (instruction length, etc.) as what the model has seen during training. Notably, the model trained on the Hard set performs very poorly (11.0\%) on the Exploration set, demonstrating that its reasonable performance on the Hard set (65.5\%) is not a good indication of it actually learning and understanding the primitives of the underlying instruction language, namely, all regular expressions. Thus, generalization to OOD instructions remains an open challenge.






\section{Conclusion}

Instruction learning has become an increasingly popular paradigm in NLP. 
Understanding the limits and capabilities of such systems is integral in continuing to improve their performance. 
While several studies have employed instruction learning in the natural language setting, we are the first to study instruction learning in a fully synthetic environment. 
By avoiding many of the hard-to-control aspects of natural language we are able to perform controlled experiments to discover what makes instruction learning hard for transformers.

We find several attributes of instances within our dataset that make instruction learning hard in our setting and use our findings speculate about the general setting. We summarize here: 
\begin{ite}
\item Non-starfree languages are hard, suggesting that following instructions for tasks requiring modeling periodicity (such as telling whether something is even or odd) are hard for transformers.
\item Large languages are hard, suggesting that more general (less precise) instructions are hard for transformers.
\item Highly composed RegExs are hard, indicating that models struggle to interpret deeply nested instructions.
\item Instances with high execution states are hard, suggesting that \added{it is harder to correctly execute instructions when doing so requires tracking a longer context of prior steps.}
\end{ite}
We do not test these hypotheses in a natural language setting and leave this as a suggestion for future work.

As a challenge for future research, we construct a hard version of our dataset using attributes we identify as contributing to hardness. 
Since not all of the RegExs in our hard set prove difficult for our model (which gets perfect accuracy on 15\% of the test RegExs) there may be more attributes not considered in this study that predict difficulty.
On the other hand, our baseline model leaves room for improvement on this synthetic instruction learning benchmark, getting less than 90\% accuracy on over 34\% of RegExs. We offer our dataset as a challenge to help the community progress in building reliable instruction learning systems. 

\bibliography{sources}
\bibliographystyle{acl_natbib}

\clearpage
\appendix

\section{Appendix}

\subsection{Regular Languages and Expressions}
\label{app:rlangs}

Following standard definitions from formal language theory \cite{harrison1978introduction}, we define a language as a set of strings of symbols from some alphabet $\Sigma$.
The regular languages over an alphabet $\Sigma$ are defined as follows:
\begin{ite}
    \item The empty set $\emptyset$ is a regular language.
    \item For each symbol $\sigma\in\Sigma$, the singleton set $\{\sigma\}$ is a regular language.
    \item The singleton set $\{\varepsilon\}$ is a regular language, where $\varepsilon$ is the empty string.
    \item If $A$ and $B$ are regular languages, then their union $A\cup B$ is a regular language.
    \item If $A$ and $B$ are regular languages, then the set $\{ab\mid a\in A, b\in B\}$ is a regular language. This new set is called the \emph{concatenation} of $A$ and $B$ and is denoted $A\cdot B$
    \item If $A$ is a regular language, then the set $\{\varepsilon\}\cup A\cup AA\cup AAA\cup \ldots$ is a regular language. This new set is called the \emph{Kleene star} of $A$ and is denoted $A^*$.
\end{ite}

A regular \emph{expression} (RegEx) is a specification of a regular language. We denote the language specified or ``expressed'' by a RegEx $r$ as $L_r$.
In a RegEx, a symbol from $\Sigma$ or the empty string represents its own singleton set, \textit{e.g.} RegEx \regex{a} represents the language $\{\str{a}\}$.
Given two RegExs $r$ and $s$, the RegEx $r|s$ represents $L_r\cup L_s$, the union of the languages expressed by $r$ and $s$. Likewise, $rs$ represents the concatenation of the two languages, and $r\regex{*}$ represents the Kleene star of $L_r$. Parentheses are used to indicate order of operations, \textit{e.g.} \regex{(a|bab)b{*}}. We refer to union, concatenation, and Kleene star as \emph{compositional operators}.

Additionally, it has been shown that regular languages are closed under set complementation, \textit{e.g.} if $L$ is a regular language, then the set $\{\sigma\in\Sigma^*\mid\sigma\notin L\}$ (denoted $L^c$) is a regular language. It follows that the complement of a any regular language can be expressed without a dedicated complement operator, however the RegEx may be verbose \textit{e.g.} \regex{(a|b){*}((b(a|b))|((a|b)b))(a|b){*}} expresses $L_{\regex{a{*}|b}}^c$.

\subsection{Sampling RegExs}\label{sec:sample}

Algorithm~\ref{alg:sample} details our method for sampling regular expressions. In summary, we find a RegEx with the minimum number operators to express each language expressible with up to $D$ operators, then sample from this set uniformly at random with respect to number of operators.

\begin{algorithm}
\caption{Sampling $N$ RegExs with at most $D$ operators. $R_d$ is the set of RegExs with $d$ compositions.}\label{alg:sample}
\begin{algorithmic}
\Procedure{Sample}{$D,N$}
\State $L\gets\emptyset$
\State $S\gets\emptyset$
\For{$d\in0,1,\ldots,D$}
\For{$r\in R_d$}
\If{$L_r\notin L$}
\State $S\gets S\cup\{r\}$
\State $L\gets L\cup\{L_r\}$
\EndIf
\EndFor
\EndFor
\State $T\gets\emptyset$
\For{$d\in0,1,\ldots,D$}
\State $n\gets\min\left(\frac N{D-d},|S\cap R_d|\right)$
\State $r_1,r_2,\ldots,r_n\sim\text{Unif}(S\cap R_d)$
\State $T\gets T\cup\{r_1,r_2,\ldots,r_n\}$
\EndFor
\State \textbf{return} $T$
\EndProcedure
\end{algorithmic}
\end{algorithm}

\end{document}